\documentclass{article}

\usepackage[preprint, nonatbib]{neurips_2022}

\usepackage[utf8]{inputenc} 
\usepackage[T1]{fontenc}    
\usepackage{hyperref}       
\usepackage{url}            
\usepackage{booktabs}       
\usepackage{amsfonts}       
\usepackage{nicefrac}       
\usepackage{microtype}      

\usepackage{amsmath,amsfonts,bm}

\def\eqref#1{equation~\ref{#1}}

\def\1{\bm{1}}

\DeclareMathAlphabet{\mathsfit}{\encodingdefault}{\sfdefault}{m}{sl}
\SetMathAlphabet{\mathsfit}{bold}{\encodingdefault}{\sfdefault}{bx}{n}

\DeclareMathOperator*{\argmin}{arg\,min}

\usepackage{amssymb} 

\usepackage[ruled,vlined,linesnumbered,noresetcount]{algorithm2e}
\SetKwInput{KwInput}{Input} 
\SetKwInput{KwOutput}{Output} 

\usepackage{enumitem}

\usepackage{bbm}

\usepackage{mathtools}

\usepackage[usenames, dvipsnames]{xcolor}
\hypersetup{colorlinks,citecolor=NavyBlue,
linkcolor=NavyBlue,urlcolor=NavyBlue}
\usepackage[nameinlink,capitalise]{cleveref}

\newcommand{\Tau}{\mathcal{T}}

\usepackage{wrapfig}

\usepackage{circuitikz}
\usetikzlibrary{patterns,hobby,decorations.pathmorphing}

\usepackage{xfp}

\usepackage{caption}
\usepackage{subcaption}

\newcommand{\model}[1]{{\fontfamily{lmtt}\selectfont {#1}}}
\newcommand{\ourmethod}{Meta-SysId}

\usepackage[backend=bibtex,natbib=true,style=ieee,mincitenames=1,maxcitenames=2,maxbibnames=20,url=false,doi=false]{biblatex}
\AtEveryBibitem{
    \ifentrytype{inproceedings}{
         \clearlist{address}
         \clearlist{publisher}
         \clearname{editor}
         \clearlist{organization}
         \clearfield{url}  
         \clearfield{doi}  
         \clearfield{pages}  
         \clearlist{location}
         \clearlist{month}
     }{}
     \ifentrytype{article}{
         \clearlist{address}
         \clearlist{publisher}
         \clearname{editor}
         \clearlist{organization}
         \clearfield{url}  
         \clearfield{doi}  
         \clearlist{location}
     }{}
 }
\usepackage{tikz}
\pgfdeclarelayer{bg}    
\pgfsetlayers{bg ,main}

\usepackage{siunitx}
\title{\ourmethod: A Meta-Learning Approach for Simultaneous Identification and Prediction}

\author{
    Junyoung Park \\
    KAIST \\ 
    \texttt{Junyoungpark@kaist.ac.kr} \\
    \And
    Federico Berto \\
    KAIST \\ 
    \texttt{fberto@kaist.ac.kr} \\
    \And
    Arec Jamgochian \\
    Stanford University\\ 
    \texttt{arec@stanford.edu} \\
    \And
    Mykel J. Kochenderfer \\
    Stanford University\\ 
    \texttt{mykel@stanford.edu} \\
    \And
    Jinkyoo Park \\
    KAIST \\
    \texttt{jinkyoo.park@kaist.ac.kr}
}

\begin{document}

\maketitle

\begin{abstract}

In this paper, we propose Meta-SysId, a meta-learning approach to model sets of systems that have behavior governed by common but unknown laws and that differentiate themselves by their context. Inspired by classical modeling-and-identification approaches, Meta-SysId learns to represent the common law through shared parameters and relies on online optimization to compute system-specific context. Compared to optimization-based meta-learning methods, the separation between class parameters and context variables reduces the computational burden while allowing batch computations and a simple training scheme. We test Meta-SysId on polynomial regression, time-series prediction, model-based control, and real-world traffic prediction domains, empirically finding it outperforms or is competitive with meta-learning baselines.

\end{abstract}

\section{Introduction}
\label{section:intro}

Natural and engineered systems are often described by parameterizing a mathematical model and finding proper system parameters within that model class \cite{ljung2010perspectives}. This \textit{modeling and system identification} paradigm has made remarkable advances in modern science and engineering \cite{schrodinger1926undulatory, black1973pricing, hawking1975particle}. However, large datasets and recent advances in deep learning tools have made it possible to model a target system without explicit knowledge, relying instead on overly flexible model classes \cite{jumper2021highly, brunton2016discovering, degrave2022magnetic, kochkov2021machine}.

However, natural and engineered systems often change their behavior for various reasons, necessitating quick model adaptation with little data. One possible approach for addressing this issue is through meta-learning, which learns a meta-model $f_\theta(\cdot)$ that can quickly adapt to a new target system.
Meta-learning methods can be classified into optimization-based and black-box methods depending on their adaptation mechanism \cite{hospedales2020meta}. Optimization-based methods explicitly optimize model parameters $\theta$ to find the task-adapted parameter $\theta'$ of adapted model $f_{\theta'}(\cdot)$. 
On the other hand, black-box methods use a context-dependent prediction model $f_\theta(\cdot \, ;c)$ and employ an inference model to extract task-specific context $c$ from new data. 

In this study, we aim to model sets of systems whose behaviors are governed by common but unknown laws and where individual systems are differentiated by their context. We propose \ourmethod, a meta-learning method to model the target system. Inspired by the modeling and identification framework, \ourmethod{} considers target systems of the form $y=f_{\theta}(x; c)$, where $\theta$ denotes the function class shared by the target systems and $c$ denotes system-specific context characteristics. \ourmethod{} learns the function class $f_\theta$ during meta-training and solves the system identification problem through numerical optimization to find the proper system-specific context $c$. \ourmethod{} can be seen as a hybridization of black-box and optimization-based meta-learning. To consider our target problem, it separates the roles of $c$ and $\theta$ as done by black-box approaches. However, like optimization-based approaches, the adaptation of \ourmethod{} is done through online optimization.

Moreover, the separation between $\theta$ and $c$ provides two major advantages over other optimization-based meta-learning algorithms (e.g., \cite{finn2017model,nichol2018first}). First, during testing, this separation reduces the burden of adaptive mechanisms, which only optimize for task-specific context $c$ while fixing class parameters $\theta$. In addition, this separation allows us to employ a training trick based on the exponential moving average (EMA), enabling us to train \ourmethod{} with gradient descent methods, and critically without computing second-order gradients. The contributions of this work are summarized as follows:
\begin{itemize}[leftmargin=0.5cm]
\vspace{-0.25cm}
    \item We propose \ourmethod, a meta-learning approach hybridizing optimization-based and black-box methods that effectively captures shared model class information and enables online adaptability through optimization.
    \item We propose a simple yet effective trick for training \ourmethod{} that overcomes the burden of second-order derivative calculations.
    \item We empirically demonstrate that \ourmethod{} outperforms or is competitive to various meta-learning baselines in static function regression, time-series prediction, model-based control, and real-world traffic flow prediction domains.
\end{itemize}


\section{Related works}
\begin{figure}[t]
     \centering
     \begin{subfigure}[b]{0.325\textwidth}
         \centering
         \includegraphics[width=\textwidth]{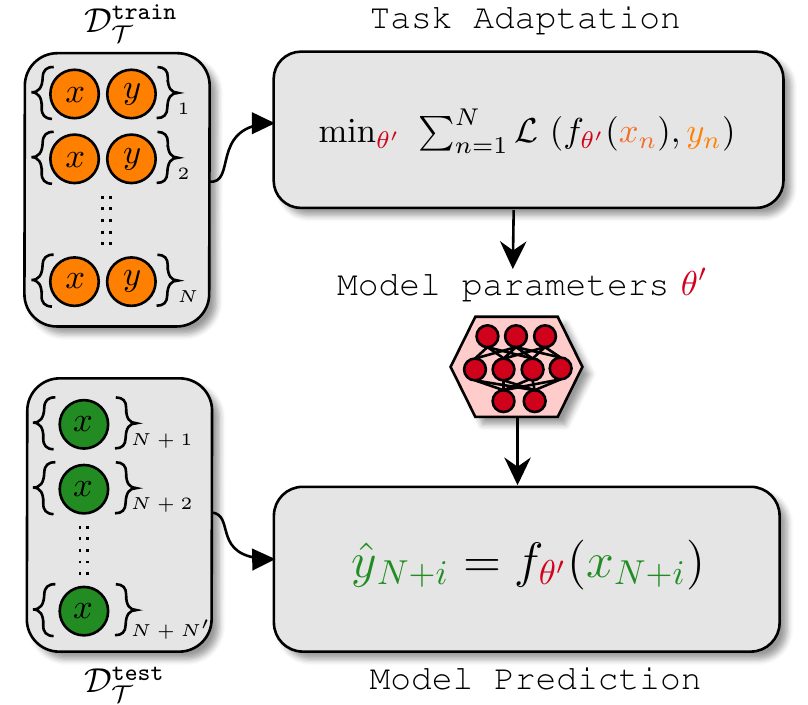}
         \caption{Optimization-based}
     \end{subfigure}
     \hfill
     \begin{subfigure}[b]{0.325\textwidth}
         \centering
         \includegraphics[width=\textwidth]{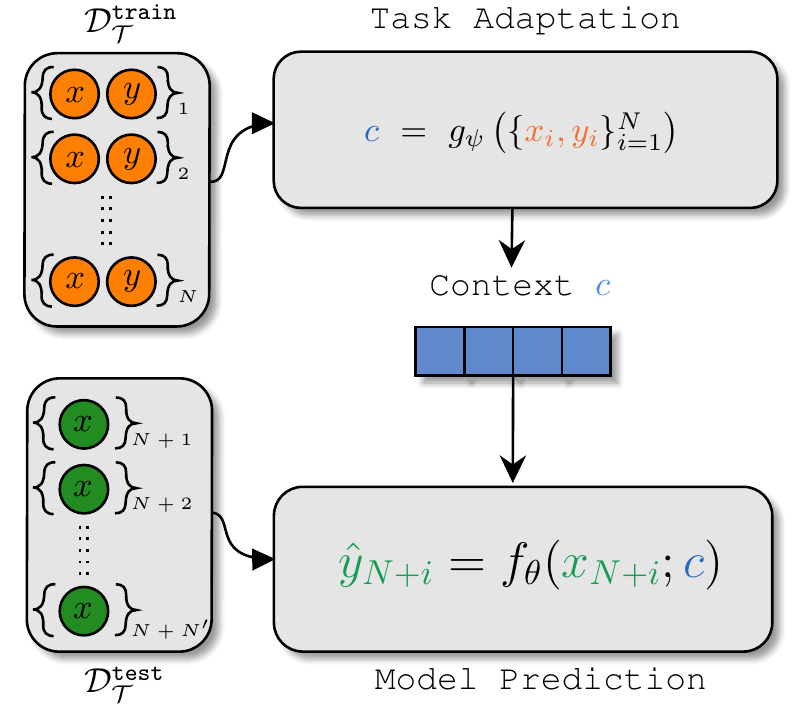}
         \caption{Black-box}
     \end{subfigure}
     \hfill
          \begin{subfigure}[b]{0.325\textwidth}
         \centering
         \includegraphics[width=\textwidth]{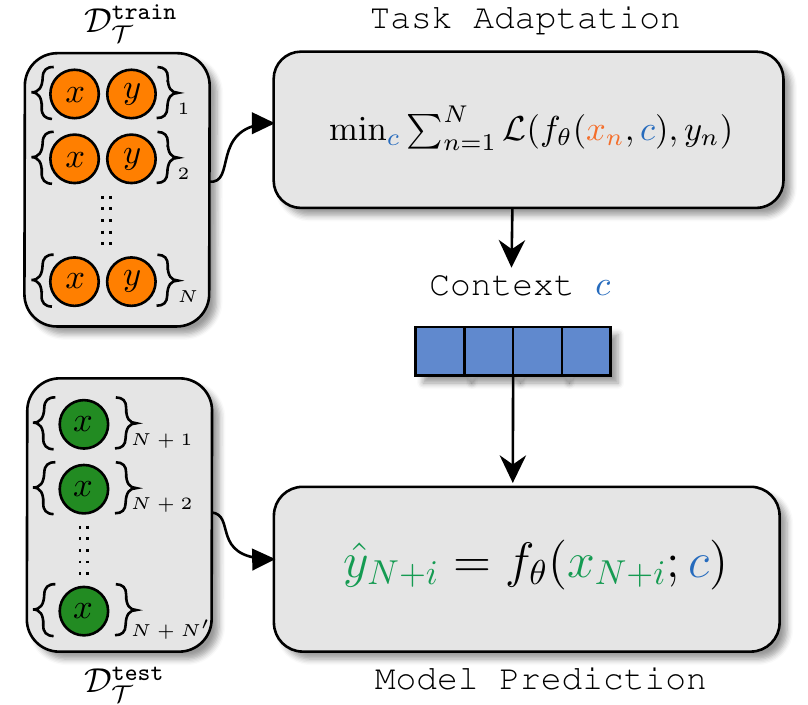}
         \caption{\ourmethod{}}
     \end{subfigure}
    \caption{\footnotesize Differences in meta--training and meta--testing between different meta-learning approaches. (a) Optimization--based models do not explicitly divide the parameters and context information, but optimize all model parameters online. (b) Black--box models obtain context information via the encoder network. (c) \ourmethod{} obtains context information by solving an optimization problem while preserving the system class information.}
    \label{fig:catchy}
\end{figure}

In this section, we detail optimization-based and black-box meta-learning approaches and discuss the unique aspect of \ourmethod{} to solve our targeted meta-learning problems. \cref{fig:catchy} illustrates the differences in the  optimization-based and black-box approaches, and \ourmethod{}.

\paragraph{Optimization-based meta-learning}
Optimization-based meta-learning methods (e.g., MAML \cite{finn2017model}, Reptile \cite{nichol2018first}) perform online optimization to find task-adapted model parameters. 
These approaches typically formulate the meta-learning task as bi-level optimization to find meta-parameters while considering the task-specific parameter adaptations. Different approaches solve the bi-level optimization by (1) formulating the inner optimization so that it has a closed-form solution \cite{bertinetto2018meta}, (2) differentiating the argmin via implicit differentiation \cite{rajeswaran2019meta, lee2019meta}, or (3) backpropagating through the inner optimization step \cite{finn2017model}. The first two methods limit the model class into a certain family to estimate the gradient without bias \cite{blondel2021efficient, liao2018reviving}, while the third method entails intensive memory usage during training and possibly biases the solution of the inner-level optimization \cite{antoniou2018train, ye2021train}. The separation of $\theta$ and $c$ in \ourmethod{} allows us to use the EMA trick for training, enabling us to use arbitrary network architectures (i.e., model agnosticism) while avoiding backpropagating through the inner optimization steps.

\paragraph{Black-box meta learning}
Black-box meta-learning methods (e.g., MANN \cite{santoro2016meta}, SNAIL \cite{mishra2018simple}, Neural Processes \cite{garnelo2018conditional, gordon2019convolutional, kim2019attentive}) jointly meta-train an inference network $g_\psi(\cdot)$ alongside the prediction model $f_\theta(\cdot;\cdot)$. 
The task-adapted predictions $y=f_\theta(x;c)$ are made using context $c$ extracted through the inference model $g_\psi$. 
Black-box methods are easier to implement than optimization-based meta-learning methods and effectively capture the shared structure between tasks through $\theta$. For our target problems, the role of $g_\psi$ is to generate a solution for classical system identification. Hence, $g_\psi$ can be viewed as an amortized optimization solver \cite{amos2022tutorial}. The adaptation capability of black-box methods can therefore be limited by the representation capacity of $g_\psi$ (typically called an amortization gap). \ourmethod{} also uses $c$ to adapt, but by employing online optimization to find $c$, it bypasses the amortization gap.

\section{Preliminaries}
\label{section:preliminary}
This section reviews classical parameter identification and meta-learning as preliminaries for explaining the proposed method.

\paragraph{Classical parameter identification}
Consider the response $y$ from an input $x$ of a system $f$ with an unknown (or unobservable) context $c$,
\begin{align}
\label{eqn:contexualized-system}
y=f(x;c)\text{.}
\end{align}

Classical parameter identification aims to infer the (optimal) context $c^*$ with data from task $\Tau=\{(x_n,y_n)\}_{n=1}^{N}$. Under the assumption that $(x_n,y_n)$ pairs in $\Tau$ are generated from the same context, the identification process can be formulated as an optimization problem as follows:
\begin{align}
\label{eqn:classical-pid}
c^*= \argmin \sum_{n=1}^{N}\mathcal{L}(f(x_n;c),y_n)\text{,}
\end{align}
where $\mathcal{L}$ is a discrepancy metric. By using the optimized context $c^*$, we can query responses of the (identified) system $f(\cdot ; c^*)$ to inputs. 


\paragraph{Meta-learning}
Meta-learning is a collection of algorithms that learn to adapt quickly to a new task by leveraging the learning experiences from related tasks. Assuming that tasks are drawn from a distribution $p(\mathcal{T})$ and yield training data $\mathcal{D}_{\mathcal{T}}^{\mathrm{tr}}$ and testing data $\mathcal{D}_{\mathcal{T}}^{\mathrm{test}}$, the meta-learning objective can be defined as follows:
\begin{align}
\label{eqn:meta-learning-obj}
& \min_{\theta, \psi} \mathbb{E}_{\Tau \sim p(\Tau)}
\left[\mathcal{L}\left(\mathcal{D}_{\Tau}^{\text{test}}, f_{\theta^{\prime}}(\cdot)\right)\right] \\
& \text{ s.t. } \quad \theta^{\prime}=g_{\psi}\left(\mathcal{D}_{\Tau}^{\mathrm{tr}}, \theta\right)\text{,}
\end{align}
where $g_\psi(\cdot, \theta)$ is a (meta)-learned adaptation algorithm that is parameterized by $\theta$ and $\psi$, $f_{\theta^{\prime}}$ is the task-specific model that is parameterized by the adapted parameter $\theta^{\prime}$, and $\mathcal{L}$ is a loss metric. 

In summary, meta-learning uses $\mathcal{D}_{\mathcal{T}}^{\mathrm{tr}}$ to learn $\theta$ and $\psi$ such that they minimize the generalization error on $\mathcal{D}_{\mathcal{T}}^{\text{test}}$. Optimized meta-parameters $\theta^*$  and $\psi^*$ can then be used to ``quickly'' adapt to new tasks from $p(\Tau)$.

\section{Meta system identification}

In this section, we present meta system identification (\ourmethod{}) to learn to jointly identify and predict systems in the same family. We first introduce the training problem of \ourmethod{} and then discuss the technical details for efficiently solving the training problem.

\subsection{Problem formulation}

\begin{figure}[t]
    \begin{subfigure}[t]{0.24\textwidth}
        \centering
        \begin{tikzpicture}[remember picture, every node/.style={inner sep=0,outer sep=0}]
            \node[anchor=south west,inner sep=0] (imageA) at (0,0) {\includegraphics[width=1.0\textwidth]{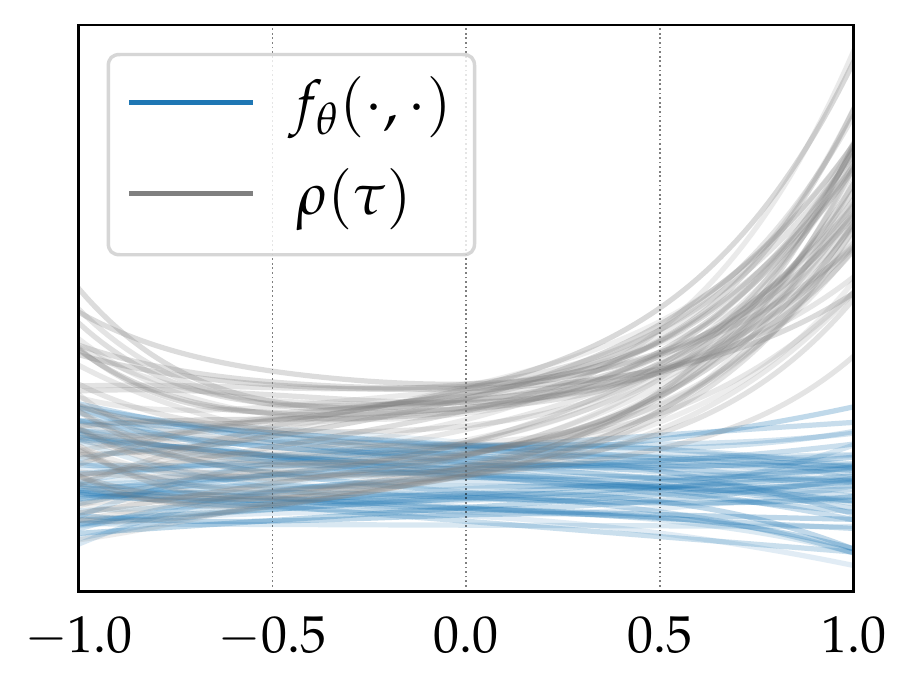}};
            \begin{scope}[x={(imageA.south east)},y={(imageA.north west)}]
                \node[coordinate] (A_right) at (1.1, .5) {};
            \end{scope}
        \end{tikzpicture}
        \caption{Initial $f_\theta(\cdot; \cdot)$}
    \end{subfigure}
    \hfill
    \begin{subfigure}[t]{0.24\textwidth}
        \centering
        \begin{tikzpicture}[remember picture, every node/.style={inner sep=0,outer sep=0}]
            \node[anchor=south west,inner sep=0] (imageB) at (0,0) {\includegraphics[width=1.0\textwidth]{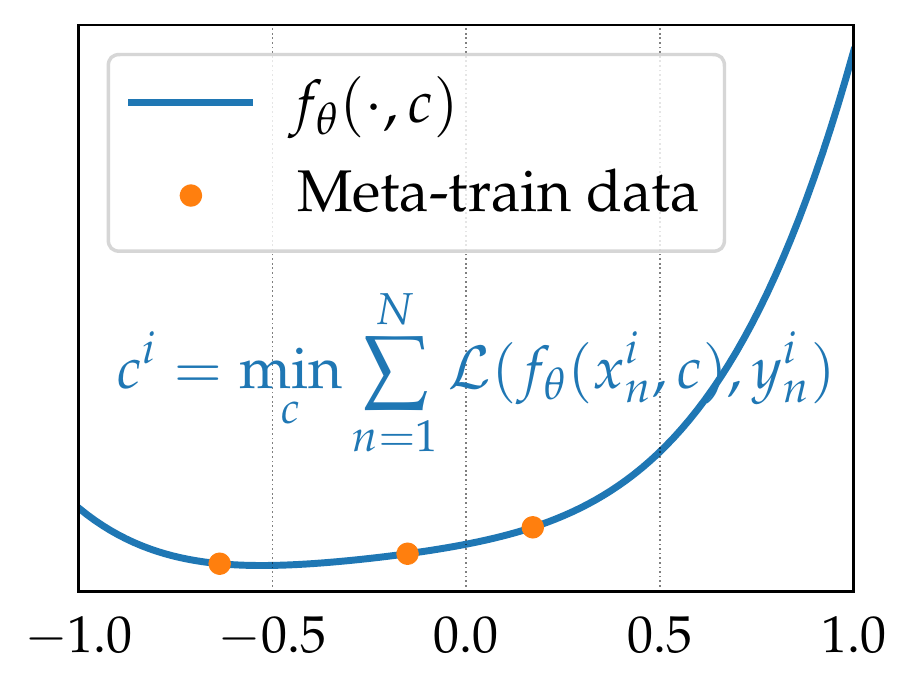}\label{fig:f_conditioned_c}};
            \begin{scope}[x={(imageB.south east)},y={(imageB.north west)}]
                \node[coordinate] (B_left) at (-0.1,0.5) {};
                \node[coordinate] (B_top) at (0.5,1.02) {};
                \node[coordinate] (B_right) at (1.1,0.5) {};
                \node[coordinate] (BC) at (1.05, 1.26) {};
                \node[coordinate] (B_botleft) at (-0.01, -0.025) {};
            \end{scope}
        \end{tikzpicture}
        \caption{Task adaptation}
    \end{subfigure}
    \hfill
    \begin{subfigure}[t]{0.24\textwidth}
        \centering
        \begin{tikzpicture}[remember picture, every node/.style={inner sep=0, outer sep=0}]
            \node[anchor=south west,inner sep=0] (imageC) at (0,0) {\includegraphics[width=1.0\textwidth]{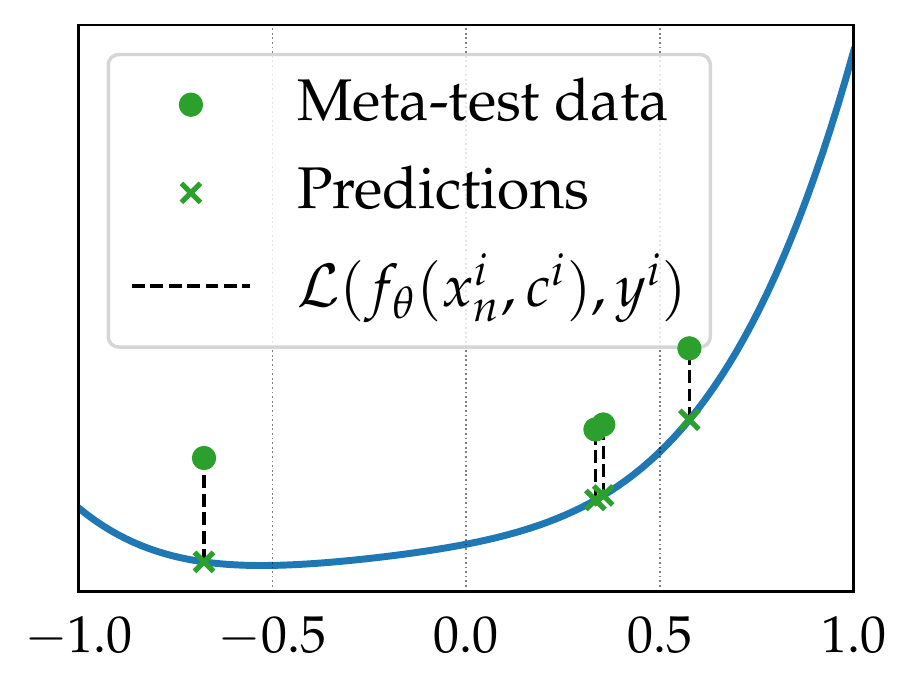}\label{fig:meta-pred}};
            \begin{scope}[x={(imageC.south east)},y={(imageC.north west)}]
                \node[coordinate] (C_left) at (-0.1, 0.5) {};
                \node[coordinate] (C_top) at (0.5, 1.02) {};
                \node[coordinate] (C_right) at (1.1, 0.5) {};
                \node[coordinate] (iter) at (0.70, 1.06) {};
                \node[coordinate] (C_topright) at (1.01, 1.17) {};
            \end{scope}
        \end{tikzpicture}
        \caption{\footnotesize Meta-prediction}
    \end{subfigure}
    \hfill
    \begin{subfigure}[t]{0.24\textwidth}
        \centering
        \begin{tikzpicture}[remember picture, every node/.style={inner sep=0, outer sep=0}]
            \node[anchor=south west,inner sep=0] (imageD) at (0,0) {\includegraphics[width=1.0\textwidth]{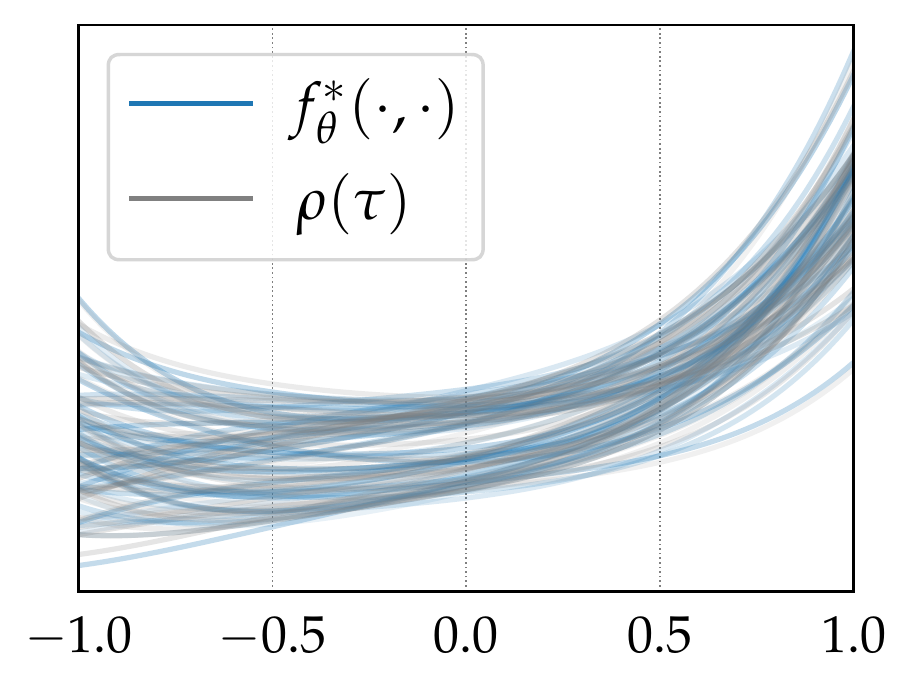}};
            \begin{scope}[x={(imageD.south east)},y={(imageD.north west)}]
                \node[coordinate] (D_left) at (-0.1, 0.5) {};
            \end{scope}
        \end{tikzpicture}
        \caption{Trained $f_\theta^*(\cdot; \cdot)$}
    \end{subfigure}
    \caption{\footnotesize \textbf{Training of \ourmethod{}} (a) the function class captured by the initial \textcolor{NavyBlue}{$f_\theta(\cdot; \cdot)$} and task distribution $\rho(\tau)$,
    (b) inference of \textcolor{NavyBlue}{$c^i$} by solving \cref{eq:meta-id-inner} with \textcolor{Orange}{meta-train data} and identification of \textcolor{NavyBlue}{$f_{\theta}(\cdot, c^i)$}, (c) \textcolor{ForestGreen}{Prediction} of $f_{\theta}(\cdot, c^i)$ and the outer loss
    $\mathcal{L}(f_\theta(x^i_{n};c^i),y^i_{n})$, (d) the optimized \textcolor{NavyBlue}{$f^*_\theta(\cdot, \cdot)$}. \ourmethod{} is trained by repeating (b), (c), and minimizing $\mathcal{L}(f_\theta(x^i_{n};c^i),y^i_{n})$.}
    \label{fig:meta_sys_id}
\begin{tikzpicture}[remember picture, overlay]
  \draw [ultra thick, black, stealth-] (B_right) to (C_left);
  \begin{pgfonlayer}{bg}
    \draw [rounded corners=5pt,
           draw=gray, ultra thick] (B_botleft) rectangle (C_topright);
  \end{pgfonlayer}
  \draw (BC) node {\textbf{Meta-training}};
  \draw (iter) node {\footnotesize $i=1,...,I$};
\end{tikzpicture}
\end{figure}
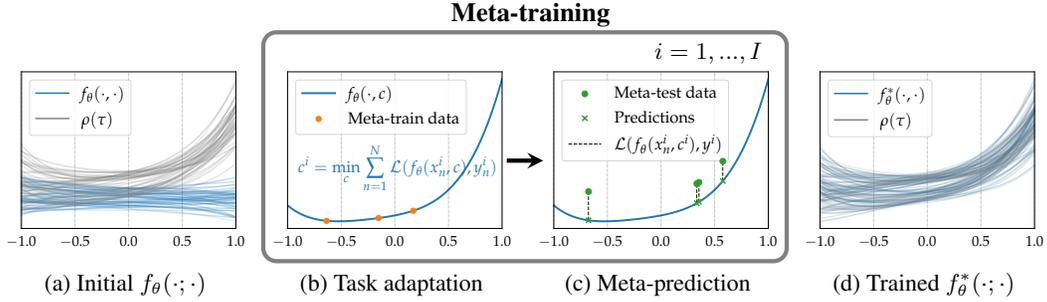

We formulate the learning problem of \ourmethod{} as a bi-level optimization where the inner optimization \cref{eq:meta-id-inner} is for identifying the context, and the outer optimization \cref{eq:meta-id-outer} is for learning the class conditioned on the identified context. The proposed optimization problem is as follows:
\begin{align}
\min_{\theta} \quad & \mathbb{E}_{\Tau^i \sim \rho(\Tau)} \sum_{n=N+1}^{N+N'}\mathcal{L}\left( f_\theta(x^i_{n};c^i),y^i_{n}\right) \label{eq:meta-id-outer}\\
\text{subject to} \quad & c^i = \argmin_c \sum_{n=1}^{N}\mathcal{L}\left( f_\theta(x^i_n;c),y^i_n \right)\text{,} \label{eq:meta-id-inner}
\end{align}
where $\Tau^i=\{(x^i_n, y^i_n)\}_{n=1}^{N+N'}$ is the $i$-th task, $N$ and $N'$ are the meta-train and -test dataset sizes, respectively, $f_\theta$ is a prediction model parameterized by $\theta$, and $\mathcal{L}$ is a loss metric. 

By solving \cref{eq:meta-id-inner,eq:meta-id-outer}, $f_\theta$ is trained. The trained $f_\theta^*(\cdot; \cdot)$ represents the function class that is shared by all tasks from $p(\Tau)$. Identification of the specific function (i.e., task adaptation) is done by solving \cref{eq:meta-id-inner} with the data points sampled from a specific task. After finding $c^i$, we can query for meta-test predictions with $f_\theta(\cdot, c^i)$ as visualized in \cref{fig:meta_sys_id} (b) and (c).



\ourmethod{} can be viewed as a special case of the meta-learning recipe explained in \cref{section:preliminary} by considering adapted parameters $\theta'$ to contain fixed class parameters $\theta$ and optimized context $c^i$. Even though \ourmethod{} is formulated similarly to optimization-based meta-learning methods, the optimization variables of \ourmethod{} are the context input $c^i$ alone rather than the model parameters. In terms of implementation, this algorithmic selection allows us to batch-solve inner-level optimization with standard automatic differentiation tools. \ourmethod{} can therefore be trained efficiently. 
We can interpret this clear separation of $\theta$ and $c$ as capturing function classes and their coefficients separately. We further inspect this view of \ourmethod{} in \cref{subsec:poly}.


\SetAlgoNoLine%

\begin{algorithm}[t]
\caption{Training \ourmethod{} with \textcolor{ForestGreen}{exponential moving average (EMA)}}
\label{alg:training-meta-id}
\DontPrintSemicolon
  \KwInput{Prediction model $f_\theta$, 
          Tasks $\mathcal{D}_{\Tau}$,
          inner optimization steps $K$,
          inner optimization step size $\alpha$,
          Weighting factor $\tau$,
          }
  $\textcolor{ForestGreen}{\bar \theta} \leftarrow \theta$, $\textcolor{ForestGreen}{f_{\bar \theta}}\leftarrow f_\theta$ \tcp*{Initialize the target model}
  \For{$1,2, ...$}
  { 
    Sample batch of tasks $\Tau^i \sim \mathcal{D}_{\Tau}$ \\
    \For{\text{all} $\Tau^i$}
    {
    $c^i \leftarrow 0$ \\
    \For{$k=1,...,K$}{
        $\mathcal{L}(c^i) = \sum_{n=1}^{N}\mathcal{L}( \textcolor{ForestGreen}{f_{\bar \theta}}(x^i_n;c),y^i_n)$ \\
        $c^i \leftarrow c^i - \alpha \nabla_{c^i}\mathcal{L}(c^i)$ \tcp*{Solve \cref{eq:meta-id-inner}}
        }
    }
  Evaluate $l(\theta)=\sum_{\Tau^i}\sum_{n=N+1}^{N'}\mathcal{L}( f_\theta(x^i_{n};c^i),y^i_{n})$ \\
  $\theta \leftarrow \theta - \alpha \nabla_\theta l(\theta)$ \\
  $\textcolor{ForestGreen}{\bar \theta} \leftarrow \tau \theta + (1-\tau) \textcolor{ForestGreen}{\bar \theta}$ \tcp*{Update the target model}
  }
\end{algorithm}
\vspace{1cm}

\subsection{Training \ourmethod}
We use a simple-yet-effective method based on EMA to train \ourmethod{} while maintaining model agnosticism and lowering memory consumption. First, we copy $\theta$ to create a delayed target model $f_{\bar\theta}$ parameterized by $\bar \theta$. We use $f_{\bar\theta}$ to infer $c^i$ of \cref{eq:meta-id-inner} through gradient descent. We then optimize $\theta$ by solving \cref{eq:meta-id-outer} with $c^i$. As $f_{\theta}$ and $f_{\bar\theta}$ are independent, this does not require differentiating the argmin operator to calculate the loss gradient. 
After a gradient update, we update $\bar \theta \leftarrow \tau \theta + (1-\tau) \bar \theta$ with $0 \leq \tau \ll 1$. The training procedure is summarized in \cref{alg:training-meta-id}. 


\paragraph{Intuition} The training problem of \ourmethod{} can be seen as a Stackelberg game where the outer optimization leads and the inner optimizations follow. In this game perspective, solving the training problem through implicit differentiation can find $\theta$ exactly. However, gradient estimation of implicit differentiation can be possibly biased when the inner problem is not solved optimally \cite{blondel2021efficient, liao2018reviving}. On the other hand, solving the proposed training problem with the EMA trick can be interpreted as formulating the training problem as a Nash game. The training trick can be interpreted as a variant of the proximal decomposition algorithm \cite{scutari2012monotone}, which is often employed to find Nash equilibria. We also empirically found that solving the training problem with the proposed trick is more stable during training and converges to better prediction models than implicit differentiation or backpropagation through the optimization steps. 
We compare the prediction results of \ourmethod{} with different training methods in \cref{subsec:poly}.

\section{Experiments}
In this section, we examine various properties of \ourmethod{} with static function regression, we apply \ourmethod{} for time-series prediction of a dynamical system, we show the efficacy of \ourmethod{} for model-based control, and we use \ourmethod{} to perform real-world traffic flow prediction. 

\subsection{Polynomial regression}
\label{subsec:poly}


We first examine polynomial function regression to understand the properties of \ourmethod. In this task, the objective is to identify polynomial coefficients and predict the identified functions. We prepare $4^{\text{th}}$ order polynomial datasets by sampling the coefficients from $\mathcal{U}(0.1,2.5)$, and set $N$ and $N'$ as 5 and 15, respectively.
The train and test datasets contain 500 and 200 polynomials, respectively. 
We provide the details of the model architecture, training scheme, and additional experimental results in \cref{appendix:poly_detail}.




\begin{wrapfigure}{r}{0.35\textwidth}
    \centering
    \includegraphics[width=\linewidth]{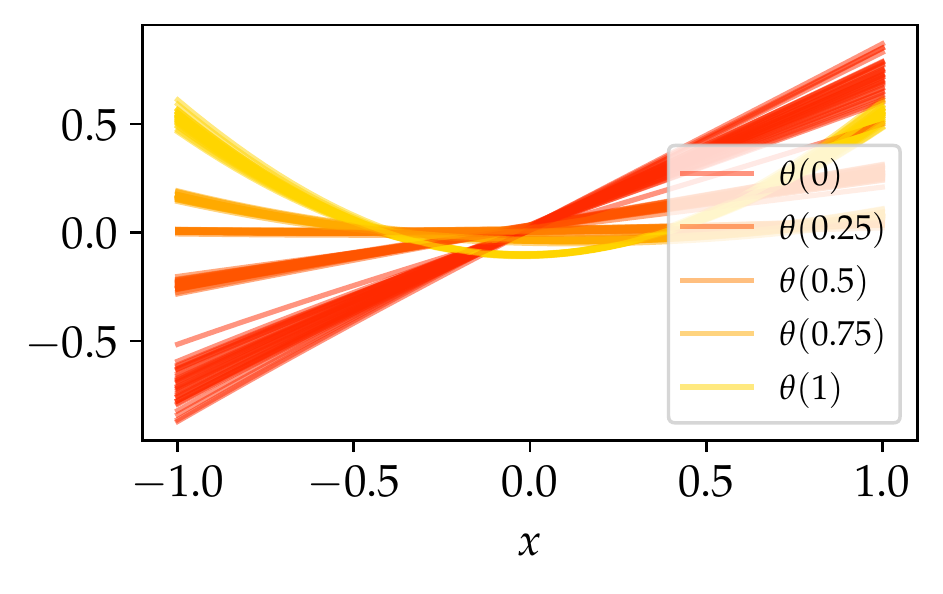}
    \caption{\footnotesize Role of varying contexts within model classes $\theta(\lambda)$ that interpolate between quadratic and linear.}
    \label{fig:poly-theta-and-c}
\end{wrapfigure}

\paragraph{Do $\theta$ and $c$ contain function class and parameter information, respectively?} The role of $\theta$ and $c$ are to capture a function class and to parameterize a specific function from that class, respectively. For example, $\theta$ should express the fact that functions are polynomials and $c$ should capture the coefficients of a specific polynomial. To analyze whether $\theta$ and $c$ fulfill these roles, we train two \ourmethod{} models: one with parameters $\theta_1$ meta-trained on linear functions, and one with parameters $\theta_2$ meta-trained on quadratic functions. We then define models that interpolate between these two sets of parameters with $\theta(\lambda) \coloneqq(1-\lambda)\theta_1+\lambda\theta_2$ with $0\leq \lambda \leq 1$. In \cref{fig:poly-theta-and-c}, we visualize predictions from $f_{\theta(\lambda)}(\cdot \, ; c)$ with different samples of  $c \sim  \mathcal{U}(-0.025,0.025)^{32}$ and different values of $\lambda$. 
We conjecture that $f_{\theta(\lambda)}(\cdot \, ; c)$ represents the linear and quadratic functions, respectively, when $\lambda=0$ and $1$ and the $f_{\theta(\lambda)}(\cdot \, ; c)$ represent the functions that interpolate the linear and quadratic functions when $0 <\lambda <1$. We verify the conjectures by observing that within a single function class $\theta(\lambda)$, each $c$ parameterizes a different function from that class (e.g., the red curves are all linear but each of them is a different linear function). 


\begin{wrapfigure}{r}{0.35\textwidth}
    \includegraphics[width=\linewidth]{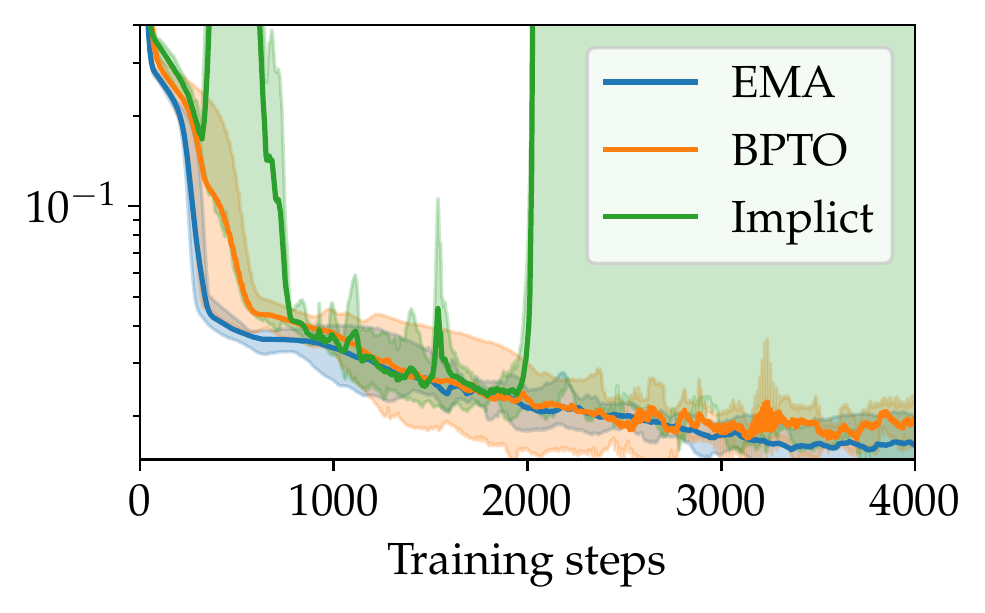}
    \caption{\footnotesize MSE vs. train steps for different training methods.}
    \label{fig:poly-training}
\end{wrapfigure}

\paragraph{Is the EMA trick an effective training scheme?} Training \ourmethod{} entails solving the bi-level training problem. To understand the effect of the training method on the performance of \ourmethod, we examine the test MSE of the different training methods. \cref{fig:poly-training} demonstrates the test MSE of \model{\ourmethod} trained with the EMA trick (\model{EMA}), backpropagation-through-optimization (\model{BPTO}), and implicit differentiation (\model{Implicit}) over the training steps. We average the results of five independent runs. From the results, we confirm that the proposed target network method exhibits stable and better training results. On the contrary, \model{Implicit} fails to converge. We attribute this difference to a biased gradient estimate as \cref{eq:meta-id-inner} generally can not be solved optimally \cite{blondel2021efficient, liao2018reviving}. It is also noteworthy that \model{EMA} has more stable and better performance than \model{BPTO}, which is trained end-to-end. This difference does not originate from the inner optimization itself since all models take the same number inner optimization steps, $K=100$. We conjecture that this happens because $f_\theta$ and $f_{\bar \theta}$ are independent during $\theta$ updates. The inferred $c^i$ acts as noise-injected context and thus training $f_\theta$ more robustly. Related work also studies the relationship between noise injection and robustness of learned models \cite{sanchez2020learning, godwin2021very, brandstetter2021message}.

\begin{wrapfigure}{r}{0.35\textwidth}
    \includegraphics[width=\linewidth]{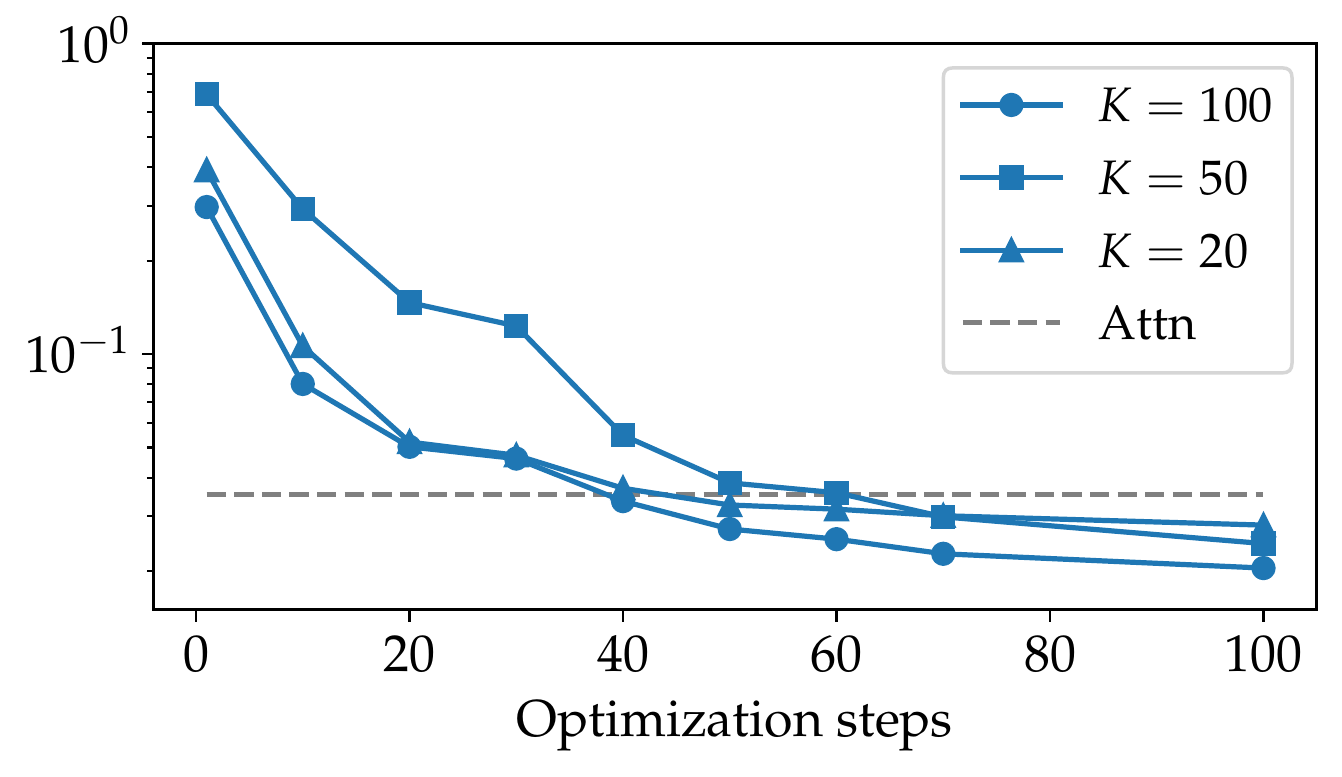}
    \caption{\footnotesize SE vs. test optimization steps for \ourmethod{} trained with the pre-specified inner optimization steps $K$.}
    \label{fig:poly-optstep}
\end{wrapfigure}

\paragraph{Does the optimization budget affect predictive performance?} The context inference of \model{\ourmethod} is done by solving \cref{eq:meta-id-inner}. As shown in \cref{alg:training-meta-id}, we employ gradient descent a pre-specified $K$ times to solve \cref{eq:meta-id-inner}. To analyze the effect of $K$ to the predictive performance of \ourmethod{}, we train \model{\ourmethod} models with different values of $K$ and evaluate them on the test dataset. \cref{fig:poly-optstep} visualizes the predictive performance of \model{\ourmethod} against the number of test time optimization steps. In general, \model{\ourmethod} with higher $K$ performs better at the cost of optimization time. We also observe that \model{\ourmethod} can improve prediction performance by spending more optimization budget when testing. Moreover, we observe that \model{\ourmethod{}} outperforms \model{Atnn} when the test optimization step is larger than 40, regardless of $K$.

\subsection{Time-series prediction}
\label{subsec:time-seires}

\begin{figure}[t]
    \centering
    \includegraphics[width=0.8\textwidth]{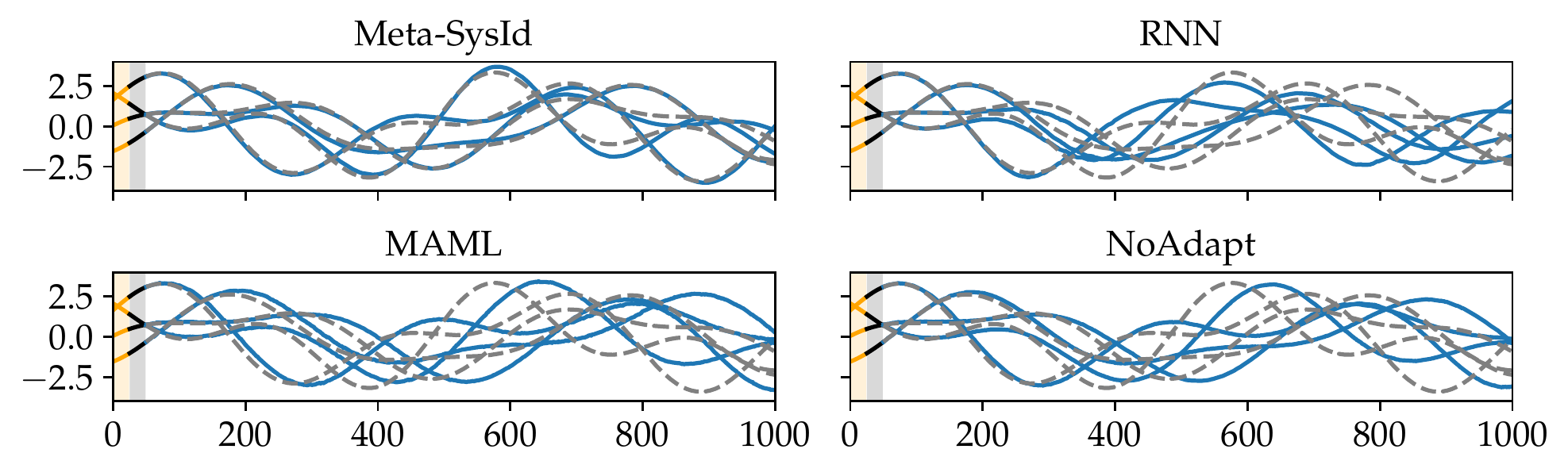}
    \caption{\footnotesize Mass-spring system prediction. All models infer the physical constants from \textcolor{YellowOrange}{historical} (shaded in \textcolor{YellowOrange}{orange}) and current (shaded in black) state observations, and then to predict \textcolor{NavyBlue}{future state trajectories} (colored in \textcolor{NavyBlue}{blue}). The \textcolor{Gray}{gray} dashed lines the visualize prediction targets. \ourmethod{} outperforms the baselines in the 25-steps and 975-steps predictions.}
    \label{fig:ts-pred-viz}
\end{figure}

We then evaluate \ourmethod{} to perform time-series prediction in mass-spring systems visualized in \cref{fig:ts-ms-system}. In this setting, the objective is first to infer the physical constants (i.e., the masses and spring constants) from historical and current state observations, and then to predict future state trajectories as shown in \cref{fig:ts-pred-viz}.


We consider the mass-spring system composed of two distinct masses and three springs without friction. We sample the masses and spring constants from ${\displaystyle {\mathcal {U}}}(0.75,1.25)$ and generate the trajectory of the four dimensional state (i.e., the positions and velocities of masses) by numerically solving the system equation for 10 seconds with 0.001 seconds intervals. Train and test datasets are composed of 100 and 50 trajectories, respectively. We set $N$ to 25 (i.e., 0.025 seconds of history is used to infer $c$). 

We employ \model{MAML} \cite{finn2017model} as an optimization-based meta-learning baseline and \model{RNN} \cite{cho2014properties} as a black-box meta-learning baseline. To understand the effect of adaptation, we employ \model{NoAdapt}, a model that does not explicitly adapt to the target tasks. For all models, including \model{\ourmethod}, we employ the same 
network for $f_\theta$, which takes $N$ recent states to predict $N$ future states. Please refer to \cref{appendix:mass_spring_detail} for architecture and training details.

\begin{figure}[t]
\captionsetup[subfigure]{justification=centering}
    \begin{subfigure}[b]{0.48\textwidth}
        \centering
        \includegraphics[width=0.95\textwidth]{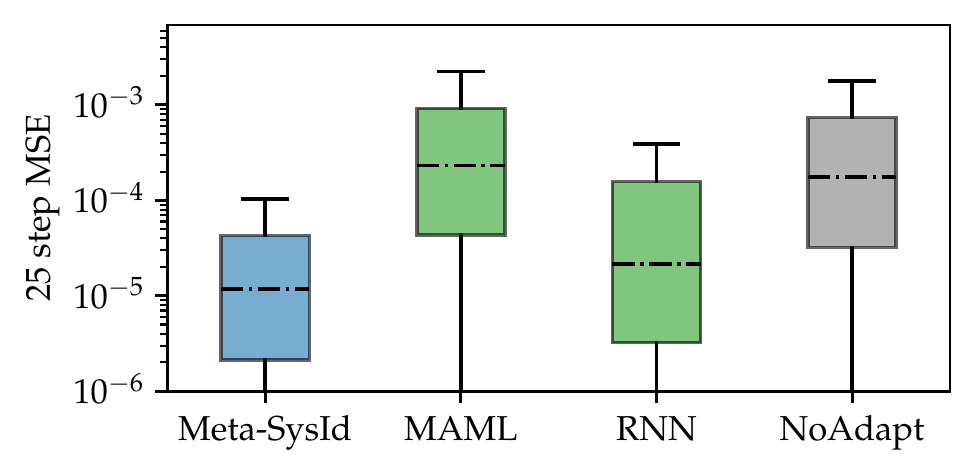}
        \vspace{-2mm}
        \caption{25-step prediction results}
        \label{fig:ts-n-step}
    \end{subfigure}
    \hfill
    \begin{subfigure}[b]{0.48\textwidth}
        \centering
        \includegraphics[width=0.95\textwidth]{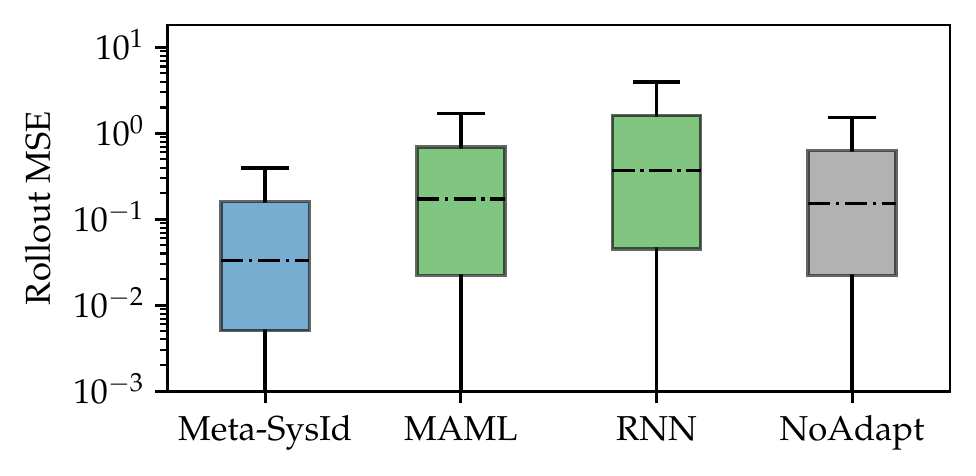}
        \vspace{-2mm}
        \caption{975-step rollout prediction results}
        \label{fig:ts-rollout}
    \end{subfigure}
    \caption{\footnotesize Mass-spring system prediction results as box-plots showing MSE quartiles. \ourmethod{}, meta-learning baselines and \model{NoAdapt} results are colored with \textcolor{NavyBlue}{blue}, \textcolor{ForestGreen}{green}, \textcolor{Gray}{gray}, respectively.}
\end{figure}

We assess the predictive performance of \model{\ourmethod} and the baseline models by evaluating MSE for 25-step predictions and 975-step rollout predictions on the test trajectories. As shown in \cref{fig:ts-n-step,fig:ts-rollout}, \model{\ourmethod} shows significantly lower $N$-step and rollout prediction errors than the baselines. Considering all models use the same $f_\theta$ architecture, we can conclude that the optimization-based context identification of \ourmethod{} helps accurately predict states both near and far into the future.


\subsection{Model-based control}
\label{subsec:control}




%
To validate \ourmethod{} is applicable to the model-based control, we consider the context-aware optimal control problem:
\begin{equation}
    \begin{aligned}
    \min_{u_{0:T-1}} \quad &  \sum_{t=0}^{T-1} J \left( x_t, u_t \right)\\
    \text{subject to} \quad & x_{t+1} = f(x_t, u_t; c_t)~ \forall t \in 0, \dots, T-1
    \label{eq:context_aware_optimal_control}
    \end{aligned}
\end{equation}

%
where the control actions $u_t$ are optimized to minimize an appropriately chosen cost function $J$. In particular, we consider the case of Model Predictive Control (MPC) \citep{camacho2013model, allgower2012nonlinear} of a Planar Fully Actuated Rotorcraft (PFAR), in which the optimization is performed \textit{online}  (see \cref{appendix:model-based-control} for additional information on the control algorithm and system dynamics). We refer to the PFAR as a ``drone'' for simplicity.

\begin{figure}
\captionsetup[subfigure]{justification=centering}
{\setlength{\fboxrule}{0pt} 
\framebox{
     \centering
     \begin{subfigure}[b]{0.18\textwidth}
         \centering
         \includegraphics[width=0.95\textwidth]{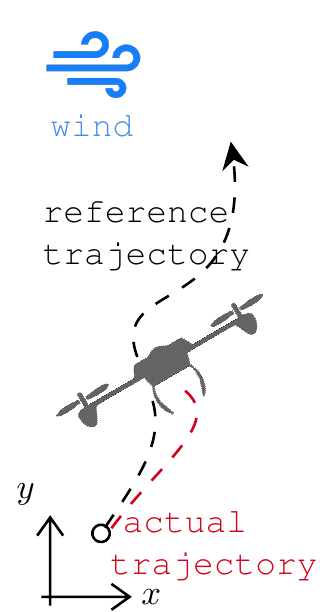}
         \caption{}
         \label{fig:reference-drone-diagram}
     \end{subfigure}
     \hfill
     \begin{subfigure}[b]{0.30\textwidth}
         \centering
         \includegraphics[width=\textwidth]{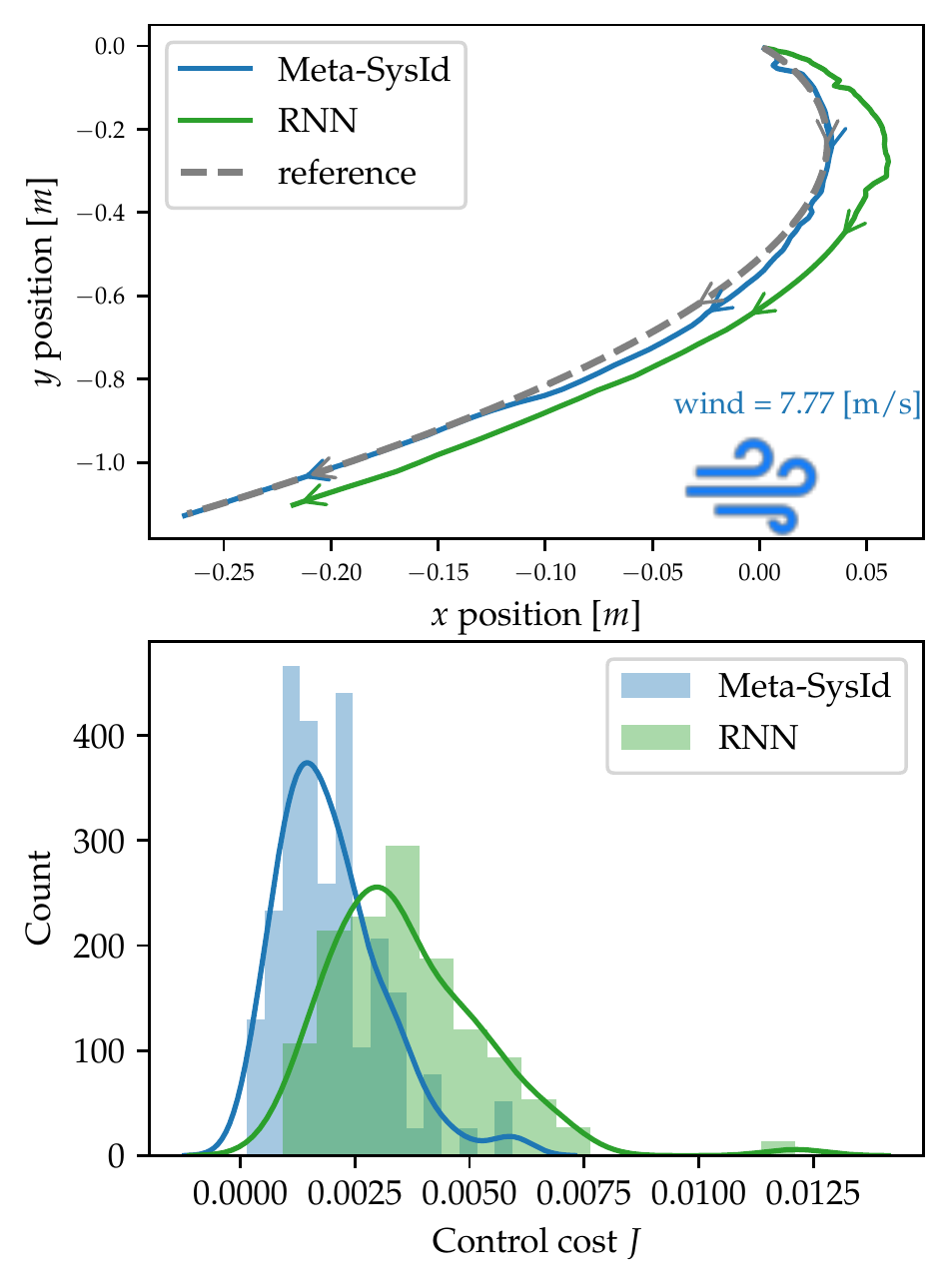}
         \caption{}
         \label{fig:reference-drone-results}
     \end{subfigure}}
     \hfill
}
{
\setlength{\fboxrule}{0pt} 
\framebox{
     \begin{subfigure}[b]{0.18\textwidth}
         \centering
         \includegraphics[width=\textwidth]{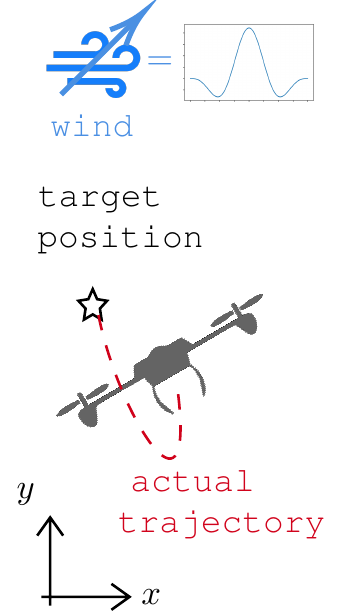}
         \caption{}
         \label{fig:stabilization-drone-diagram}
     \end{subfigure}
     \hfill
     \begin{subfigure}[b]{0.30\textwidth}
         \centering
         \includegraphics[width=0.899\textwidth]{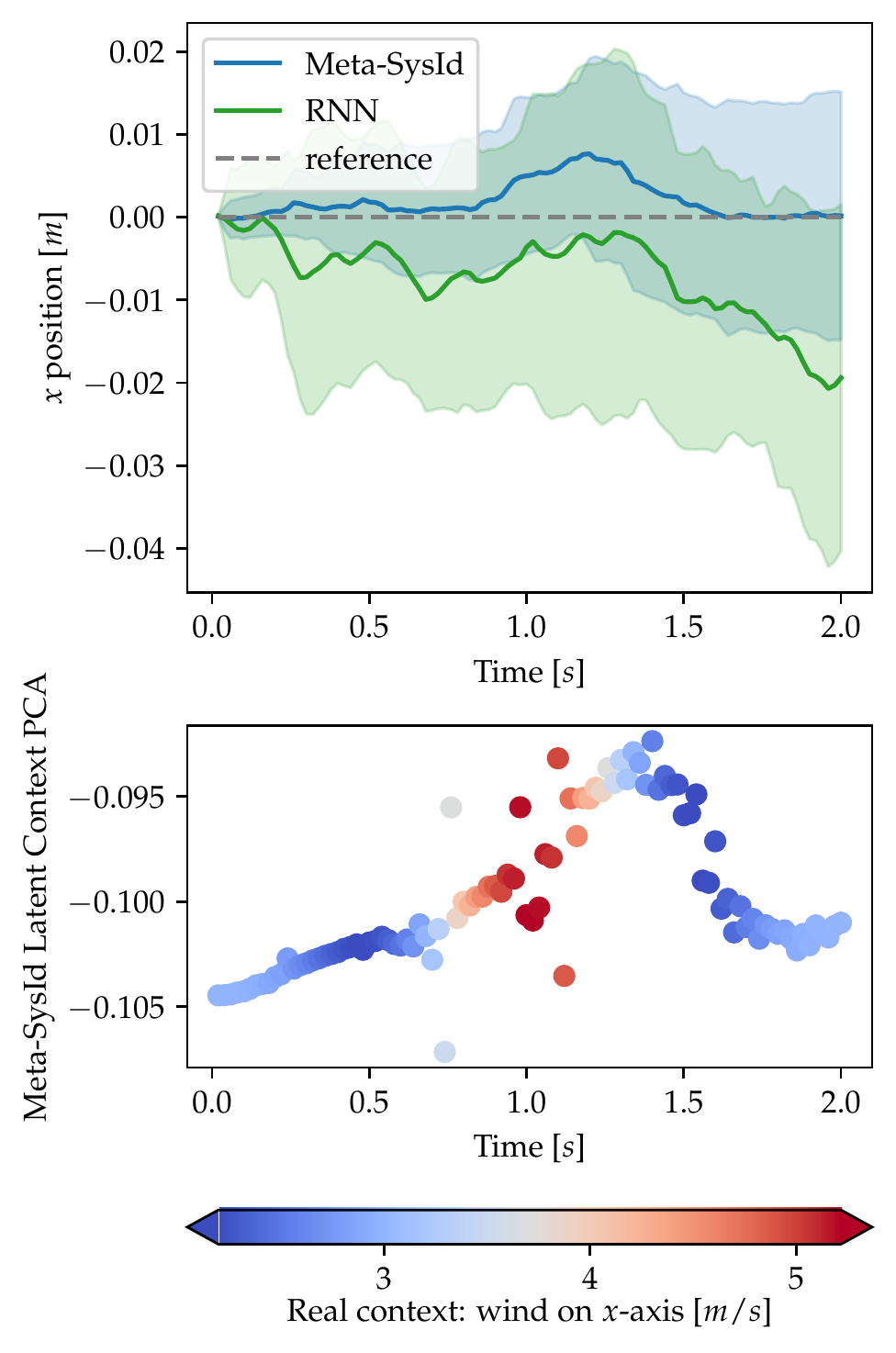}
         \caption{}
         \label{fig:stabilization-drone-results}
     \end{subfigure}}
}
\caption{\footnotesize [Left] (a) Drone trajectory tracking task diagram. (b) Sample controlled trajectory and distribution of control costs show our \model{\ourmethod{}} achieves better performance than the baseline. [Right] (c) Diagram for drone stabilization task with dynamic wind. (d) Unlike the baseline, \model{\ourmethod{}} manages to keep the drone stabilized even with changing context; the inferred context visualized by 1D PCA over time at the bottom reveals smooth changes in the real context are approximately reflected in the latent space.}
\label{fig:drone-results}
\end{figure}

We train \model{\ourmethod} and \model{RNN} \cite{cho2014properties} for $50$ epochs with the Adam optimizer \citep{kingma2014adam} with learning rate of $10^{-3}$. Context inference in \model{\ourmethod} is performed by iterating for $K=50$ steps, also with the Adam optimizer and a learning rate of $10^{-3}$. Latent context size is set to $32$ in order not to bias the model with the real context size of $1$ (horizontal wind). Models are trained on trajectories with episode-specific constant wind with a time step $\Delta t = \SI{0.02}{\second}$ and context inference on the past $25$ time steps. See \cref{appendix:drone-data-generation} for details. We introduce two separate control tasks in the following paragraphs. We also considered employing a MAML-based baseline as an optimization-based method. However, training the MAML-based models is intractable to our computing infrastructure. For more details, please refer to \cref{appendix:drone-model-training}.

\paragraph{Reference trajectory tracking with contant context} We perform MPC on the drone model to follow randomly generated smooth reference trajectories. In each episode, we apply the wind with the speed sampled from $\mathcal{U}(0,8)$ \SI{}{\meter/\second} to the $x$-axis, similar to what was done in \citep{richards2021adaptive}. Figure \cref{fig:stabilization-drone-results} shows the results of MPC for a sampled trajectory and the distribution of control costs over the 1,000 sampled trajectories, with \model{\ourmethod} achieving lower control cost than \model{RNN}. See \cref{appendix:drone-additional details} for additional details.

\paragraph{Stabilization over time--varying context} We also test the model under a time--varying context, namely a horizontal wind characterized by the profile of an \textit{extreme operating gust} \citep{branlard2009wind}. We control the drone with MPC with the task of stabilizing positions and velocities to the origin. \cref{fig:stabilization-drone-results} shows $x$-axis positions over $10$ runs and a 1--dimensional PCA visualization \citep{wold1987principal} of the latent context inferred by \model{\ourmethod} over time. While \model{RNN} suffers from trajectory drifting due to the varying context, \model{\ourmethod} manages to compensate for the gust of wind. The latent context visualization also provides a further insight: \model{\ourmethod} can infer an approximately \textit{smooth} change in latent context which matches the real--life smooth-yet-sharp change in wind. On the other hand, as shown in \cref{fig:latent-contexts-drone}, \model{RNN} does not generally yield an interpretable latent context variation.
Further details are available in \cref{appendix:drone-additional details}.

\subsection{Real-world traffic flow prediction}
\label{subsec:traffic}
\begin{table}[t]
    \centering
    \caption{PeMSD7(M) results}
    \begin{tabular}{lccc}
        \toprule  
        & MAE & MAPE & RMSE \\
        & $(15 / 30 / 45 \mathrm{~min})$ & $(15 / 30 / 45 \mathrm{~min})$ & $(15 / 30 / 45 \mathrm{~min})$ \\
        \cline{1-1} \cline{2-2} \cline{3-3} \cline{4-4}
        \model{STGCN(Cheb)} \cite{yu2017spatio} & $2.25/3.03/3.57$ & $5.26/7.33/8.69$ & $
        
        4.04/5.70/6.77$ \\
        \model{STGCN(1st)} \cite{yu2017spatio} & $2.26/3.09/3.79$ & $5.24/7.39/9.12$ & $4.07/5.77/7.03$  \\
        \model{STGCN-Cov} \cite{yoo2021conditional} & 
        $2.20/2.9 7/3.51$ & $5.14/7.26/8.74$ & $4.02/5.64/6.70$ \\
        \hline
        \model{MAML} & 
        $\textbf{2.17}/2.89/3.21$ &
        $\textbf{5.06}/6.95/7.89$ & $\textbf{3.90}/5.21/\textbf{5.81}$ \\
        \model{STGCN-Enc (black-box)}& 
        $2.24/2.85/3.54$ & $5.19/7.00/8.51$ & $3.96/5.27/6.10$ \\
        \hline
        \model{\ourmethod{}} & 
        $2.19/\textbf{2.79}/\textbf{3.16}$ &
        $\textbf{5.06}/\textbf{6.81}/\textbf{7.84}$ & $3.93/\textbf{5.17}/5.87$ \\
        \bottomrule
        \end{tabular}
    \label{table:pems}
\end{table}

As mentioned in \cref{section:intro}, \ourmethod{} is devised to solve the meta-learning tasks where each task is assumed to share a common function class and the context $c$ differentiates tasks. However, the training of \ourmethod{} can be done without these assumptions. We hypothesize that \ourmethod{} learns to generate a ``virtual" context and, by leveraging the learned context, can improve the performance of general time-series prediction tasks. To validate this hypothesis, we apply \ourmethod{} to conduct time-series prediction on a real-world traffic dataset.


We consider the PeMSD7(M) dataset \cite{yu2017spatio}, which consists of traffic measurements taken at 5-minute intervals on weekdays in May and June of 2021 at 228 locations in District 7 of California. The objective is to use 60 minutes of recent measurements to predict 15, 30, and 45 minutes into the future. We employ the spatio-temporal graph convolution network (STGCN) with the same hyperparameters \cite{yu2017spatio} for $f_\theta(\cdot)$ of \model{\ourmethod{}} with modifications to the input dimension to account for context. 
To compare against non-meta-learning approaches, we baseline against STGCN with Chebyshev polynomial approximations (\model{STGCN(Cheb)}), STGCN with first-order approximations (\model{STGCN(1st)}), and STGCN with Covariance Loss regularization (\model{STGCN-Cov}) \cite{yoo2021conditional}. For meta-learning baselines, we consider \model{MAML} \cite{finn2017model} and \model{STGCN-Enc}, a model which uses STGCN as the history encoder for adaptation (i.e., black-box approach). 
All meta-learning approaches use the last 60-minutes of observations and their corresponding labels to infer context. 
Note that we can always use this strategy to infer context inputs because they are \textit{only} composed from history. For more training details, see \cref{appendix:traffic-flow-prediction}.

\cref{table:pems} summarizes the prediction errors of the different models. We find that meta-learning approaches outperform non-meta learning baselines, but this might be the effect of using additional data (i.e., the data for adaptation). 
Among the meta-learning approaches, \model{\ourmethod{}} shows competitive or leading predictive performance, which is noteworthy because it is designed to solve meta-learning problems which separate context and function class. 
This suggests that \ourmethod{} can extract meaningful ``virtual'' context to enhance predictive performance when real context may not exist.


\section{Conclusion}
Motivated by traditional \textit{modeling and system identification} approaches, in this work, we studied how to generalize modeling for families of tasks that have unknown but shared model structures and differ in task context. We presented \ourmethod{}, an optimization-based meta-learning model that separates model class parameters from task-specific context variables. \ourmethod{} treats the context as an extra input of the prediction model and infers the context variable during meta-testing through optimization. To train \ourmethod{}, we employed an Expected Moving Average (EMA) trick to avoid differentiating through the inner context-finding optimization problem, resulting in a stable training procedure that relies exclusively on first-order gradients. We empirically found that \ourmethod{} performs competitively or outperforms optimization-based and black-box meta-learning baselines in several domains ranging from regression to online control to real-world prediction.

A current limitation of our work is that we only consider systems that can, in principle, be identifiable from the given task dataset. In the setting where tasks are sampled from nearly identical systems, \ourmethod{} may not be able to infer context properly. We expect such limitations can be overcome by augmenting the learning model with the proper selection of priors as similarly done in various inverse problem domains.

\printbibliography


\appendix
\newpage

\renewcommand{\theequation}{A.\arabic{equation}}
\renewcommand{\thetable}{A.\arabic{table}}
\renewcommand{\thefigure}{A.\arabic{figure}}
\setcounter{equation}{0}
\setcounter{table}{0}
\setcounter{figure}{0}

\section{Experiment details}
\subsection{Polynomial regression}
\label{appendix:poly_detail}
In this section, we provide the model architectures, training process, and additional experimental results for the polynomial regression task.

\paragraph{Model architecture}
We employ \model{MAML} \cite{finn2017model} as an optimization-based meta-learning baseline and a self-attention model \model{Attn} \cite{vaswani2017attention} as a black-box meta-learning baseline. All models use the same multi-layer perceptron (MLP) for $f_\theta$ and are trained with the same data batches for fair comparisons. 

For brevity, we refer to a multi-layer perceptron (MLP) with hidden dimensions $n_1$, $n_2$, ... $n_l$ for each layer and hidden activation \model{act}, as \model{MLP}($n_1$, $n_2$, ..., $n_l$; \model{act}). We refer to a cross multi-head attention block \cite{vaswani2017attention} with $h$ heads and $x$ hidden dimensions as \model{X.MHA}($h \times x$). 

\begin{table}[h!]
\caption{Polynomial regression model architectures}
\centering
\resizebox{0.95\columnwidth}{!}{
\begin{tabular}{c|c|c|c|c|c}
\toprule
      & Context encoder & $f_\theta$ & Inner step $K$ & Inner step size $\alpha$ & $\tau$\\
\midrule
\model{\ourmethod}     & $-$  & \model{MLP}(1+\textcolor{ForestGreen}{32}, 64, 32, 1;\model{SiLU} \cite{ramachandran2017searching}) & 100 & 0.001 & 0.1 \\
\model{MAML}     & $-$  & \model{MLP}(1, 64, 32, 1;\model{SiLU}) & 4 & 0.001 & $-$\\
\model{Attn}         & \model{MLP}(2,32)/\model{MLP}(1,32)-\model{X.MHA}(4$\times$\textcolor{ForestGreen}{32}) & \model{MLP}(1+\textcolor{ForestGreen}{32}, 64, 32, 1;\model{SiLU}) & $-$ & $-$ & $-$\\
\bottomrule
\end{tabular}
}
\label{table:poly-model-archi}
\end{table}


\cref{table:poly-model-archi} summarizes the network architectures. The \textcolor{ForestGreen}{green colored values} indicate the dimension of context $c$. For MAML, we also perform hyperparameter search to optimize $K$. We found that \model{MAML} with $K>5$ underperforms, as compared to $K=5$.

\paragraph{Training details}
We train all models with mini-batches of 256 polynomials for 4,048 epochs using Adam \cite{kingma2014adam} with a fixed learning rate of 0.001.



\begin{table}[h]
    \centering
    \caption{Polynomial regression MSE with increasing context points $N$.}
    \resizebox{0.5\linewidth}{!}{
    \begin{tabular}{c|ccc|c}
    \toprule
    $N$   & \ourmethod{} & \model{MAML} & \model{Attn} & \model{System. ID}  \\
    \midrule
    1     & \textbf{0.1630}  & 0.6438  & 0.3905 & 0.4989     \\
    3     & \textbf{0.0523}  & 0.0815  & 0.0826 & 0.0840     \\
    5     & \textbf{0.0268}  & 0.0473  & 0.0495 & 0.0187     \\
    10    & \textbf{0.0097}  & 0.0260  & 0.0204 & 0.0000     \\
    \bottomrule
    \end{tabular}
    }
    \label{table:poly-ctx}
\end{table}
\paragraph{Additional experiments} We evaluate the predictive performances of the models with different values of $N$ on the test dataset. \cref{table:poly-ctx} shows the average mean squared errors (MSE) of the meta-learning methods and the classical system identification that knows the exact functional form and solve the system identification problem with the context points (\model{System Id}). As shown in \cref{table:poly-ctx}, \ourmethod{} outperforms all meta-learning baselines for all $N$. Noticeably, it even shows better predictions than \model{System Id} when $N<5$.

\subsection{Time-series prediction}
\label{appendix:mass_spring_detail}

\paragraph{Mass-spring system}

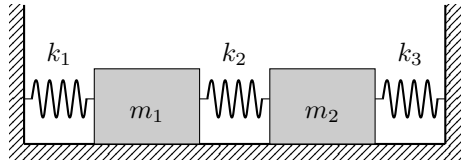
\begin{figure}[h]
    \centering
    \newcommand\wh{0.2} 
\newcommand\w{6.0} 
\newcommand\h{2.05} 

\newcommand\objc{0.8} 
\newcommand\objh{1.2} 
\newcommand\objw{1.4} 

\newcommand\sprl{\fpeval{(\w-\wh-\wh-\objw-\objw)/3}} 
\newcommand\mhh{\fpeval{(\objh)/2}} 
\newcommand\mhw{\fpeval{(\objw)/2}} 

\begin{circuitikz}

\pattern[pattern=north east lines] (0, 0) rectangle (\wh, \h);
\draw[thick] (\wh, \wh) -- (\wh, \h);

\pattern[pattern=north east lines] (\wh, 0) rectangle (\w-\wh, \wh);
\draw[thick] (\wh, \wh) -- (\w-\wh, \wh);

\pattern[pattern=north east lines] (\w-\wh, 0) rectangle (\w, \h);
\draw[thick] (\w-\wh, \wh) -- (\w-\wh, \h);

\draw (\wh, \objc) to[spring, l=$k_1$] (\wh+\sprl, \objc);

\draw[fill=gray!40] (\wh+\sprl, \wh) rectangle (\wh+\sprl+\objw, \objh);
\node at (\wh+\sprl+\mhw,\mhh) {$m_1$};

\draw (\wh+\sprl+\objw, \objc) to[spring, l=$k_2$] (\wh+\sprl+\sprl+\objw, \objc);

\draw[fill=gray!40] (\wh+\sprl+\sprl+\objw, \wh) rectangle (\wh+\sprl+\sprl+\objw+\objw, \objh);
\node at (\wh+\sprl+\sprl+\objw+\mhw, \mhh) {$m_2$};

\draw (\wh+\sprl+\sprl+\objw+\objw, \objc) to[spring, l=$k_3$] (\wh+\sprl+\sprl+\sprl+\objw+\objw, \objc);

\end{circuitikz}
    \caption{\footnotesize Target mass-spring system. The models require to adapt to the change of spring constants ($k_1, k_2, k_3$) and masses ($m_1, m_2$).}
    \label{fig:ts-ms-system}
\end{figure}

For the time-series prediction task, we consider a frictionless three-spring two-mass system shown in \cref{fig:ts-ms-system} with mass positions $x_1$ and $x_2$ governed by the following dynamics:   

\begin{equation}
    \begin{pmatrix} \Ddot{x_1}\\ \Ddot{x_2} \end{pmatrix}  =
    \underbrace{
    \begin{bmatrix}
        -\frac{k_1 + k_2}{m_1}  & \frac{k_2}{m_1} \\
        \frac{k_2}{m_2}  &  -\frac{k_2 + k_3}{m_2}  \\
    \end{bmatrix}}_{K}
    \begin{pmatrix} x_1\\ x_2 \end{pmatrix}\text{,}
    \label{eq:mass-spring-system}
\end{equation}

where $K$ is a coefficient matrix with spring constants $(k_1, k_2, k_3)$ and masses $(m_1, m_2)$.

\paragraph{Model architecture}
For $f_\theta$, we employ the 1D-CNN model from \cite{brandstetter2021message}, which stacks a MLP, 1D-CNN, and consistency decoder \cite{brandstetter2021message}. For brevity, we refer to a 1D-CNN layer with $x$ input channels, $y$ output channels, and filter size $w$ as \model{Conv}($x,y,w$), a consistency decoder as \model{C.Dec}, and the bi-directional GRU \cite{cho2014properties} with hidden dimension $x$ as \model{GRU}($x$).

\begin{table}[h!]
\caption{Mass-spring prediction model architectures}
\centering
\resizebox{1.0\columnwidth}{!}{
\begin{tabular}{c|c|c}
\toprule
      & Context encoder & $f_\theta$\\
\midrule
\model{\ourmethod}     & $-$  & \model{MLP}(4$\times$25+\textcolor{ForestGreen}{64}, 64, 32, 4$\times$25;\model{SiLU})-\model{Conv}(4, 4, 8)-\model{SiLU}-\model{Conv}(4,4,1)-\model{C.Dec} \\
\model{MAML}     & $-$  & \model{MLP}(4$\times$25, 64, 32, 4$\times$25;\model{SiLU})-\model{Conv}(4, 4, 8)-\model{SiLU}-\model{Conv}(4,4,1)-\model{C.Dec}\\
\model{RNN}         & \model{GRU}{(64)}-\model{MLP}(64, \textcolor{ForestGreen}{64}) & \model{MLP}(4$\times$25+\textcolor{ForestGreen}{64}, 64, 32, 4$\times$25;\model{SiLU})-\model{Conv}(4, 4, 8)-\model{SiLU}-\model{Conv}(4,4,1)-\model{C.Dec}\\
\model{NoAdapt} & $-$  & \model{MLP}(4$\times$25, 64, 32, 4$\times$25;\model{SiLU})-\model{Conv}(4, 4, 8)-\model{SiLU}-\model{Conv}(4,4,1)-\model{C.Dec} \\
\bottomrule
\end{tabular}
}
\label{table:ms-model-archi}
\end{table}
\cref{table:ms-model-archi} summarizes the network architectures. The \textcolor{ForestGreen}{green colored values} indicate the dimension of context $c$. We set the inner step $K$ and step size $\alpha$ as 50/5 and 0.001/0.001 for \model{\ourmethod} and \model{GrBAL}, respectively, and $\tau$ as 0.1.

\paragraph{Training details} We train all models with mini-batches of size 512 for 512 epochs using Adam \cite{kingma2014adam} with a fixed learning rate of 0.001 and pushforward regularization \cite{brandstetter2021message}. For the meta-learning models (i.e., \model{\ourmethod}, \model{MAML}, \model{ReBAL}), we use the past observations from the previous $2N$ to $N$ steps as the input of the adaptation process.

\subsection{Model-based control}
\label{appendix:model-based-control}

\subsubsection{Model predictive control formulation}
\label{appendix:model-predictive-control}

We first give a brief introduction of the control algorithm we use. Also known as receding horizon control, model predictive control (MPC) is a class of flexible control algorithms capable of taking into consideration constraints and nonlinearities \citep{camacho2013model, allgower2012nonlinear}.
MPC interleaves actions with planning over finite time windows which are then shifted forward in a \textit{receding} manner. The control problem is then solved for each window by \textit{predicting} future trajectories with a candidate controller $u$ and then adjusting the controller to optimize the cost function $J$. The first control action is taken and the optimization is repeated \textit{online} until the end of the episode.

\subsubsection{System dynamics}
\label{appendix:drone-dynamics}

We consider a Planar Fully Actuated Rotorcraft (PFAR) \citep{richards2021adaptive, richards2022control} with position $x$, height $y$, angle $\phi$, and dynamics governed by:  

\begin{equation}
    \begin{pmatrix} \Ddot{x}\\ \Ddot{y}\\ \Ddot{\phi} \end{pmatrix}  =
    \underbrace{
    \begin{bmatrix}
        \cos\phi  & -\sin\phi   & 0 \\
        \sin\phi  &  \cos\phi   & 0 \\
        0         & 0           & 1
    \end{bmatrix}}_{R(\phi)}
    \begin{pmatrix}
            u_x - \beta_1 v_1|v_1| \\ 
            u_y - \beta_2 v_2|v_2| \\
            u_\phi
    \end{pmatrix} + \begin{pmatrix}
            0 \\
            -g \\
            0
    \end{pmatrix}\text{,}
    \label{eq:drone-system}
\end{equation}

where $R(\phi)$ is a rotation matrix, $u = (u_x, u_y, u_\phi)$ are normalized control actions on their respective dimensions (i.e. thrusts and a torque), $g = 9.81 ~m/s^2$ is the gravitational acceleration, and $\beta_1$ and $\beta_2$ are drag coefficients which we set to $0.1$ and $1$ respectively. Velocities $v_1$ and $v_2$ can be described by:

\begin{equation}
\begin{aligned}
    v_1 &= (\dot{x} - w)\cos\phi + \dot{y}\sin\phi, \\
    v_2 &= -(\dot{x} - w)\sin\phi + \dot{y}\cos\phi, \\
\end{aligned}
\end{equation}

where $w$ is wind acting on the $x$ direction.

\subsubsection{Data generation details}
\label{appendix:drone-data-generation}

We generate a total of $500$, $10$ seconds--long, reference trajectories by fitting a smooth polynomial spline trajectory on a random walk of the state parameters and simulating the system with a PD controller with proportional and differential gain $k_p = 10$, $k_d = 0.1$ respectively, and trajectory--wise constant wind on the $x$-axis drawn from  ${\displaystyle {\mathcal {U}}}(0,8) ~m/s$. We note that the PD--controlled trajectories are not ``optimal'' as the tracking performance is not essential; we aim only at obtaining a rough distribution of state-control sequences to train the models. 
 
\subsubsection{Model training}
\label{appendix:drone-model-training}
We train \model{\ourmethod} and the \model{RNN} for $512$ epochs with the Adam optimizer \citep{kingma2014adam} with learning rate of $10^{-3}$. Context inference in \model{\ourmethod} is performed by iterating for $50$ steps, also with the Adam optimizer and a learning rate of $10^{-3}$. the latent context dimension is set to $32$. Models are trained on the dataset described in \cref{appendix:drone-data-generation} with the sub-sampled training trajectories. Each sub-trajectory comprises 25-time steps of historical observations and 25-time steps of future observation. All models utilize historical observations and future observations for context inference and meta prediction.

We also tested the optimization-based meta-learning approach (i.e., MAML) to the specified training scheme. However, using our computing infrastructures, training of the optimization-based meta-learning approach took more than an hour for a single epoch. As all models are trained for $512$ epochs, we exclude the MAML-based approach from the baseline for fair comparisons.



\subsubsection{Additional experimental details}
\label{appendix:drone-additional details}


\begin{figure}
    \centering
    \includegraphics[width=0.5\linewidth]{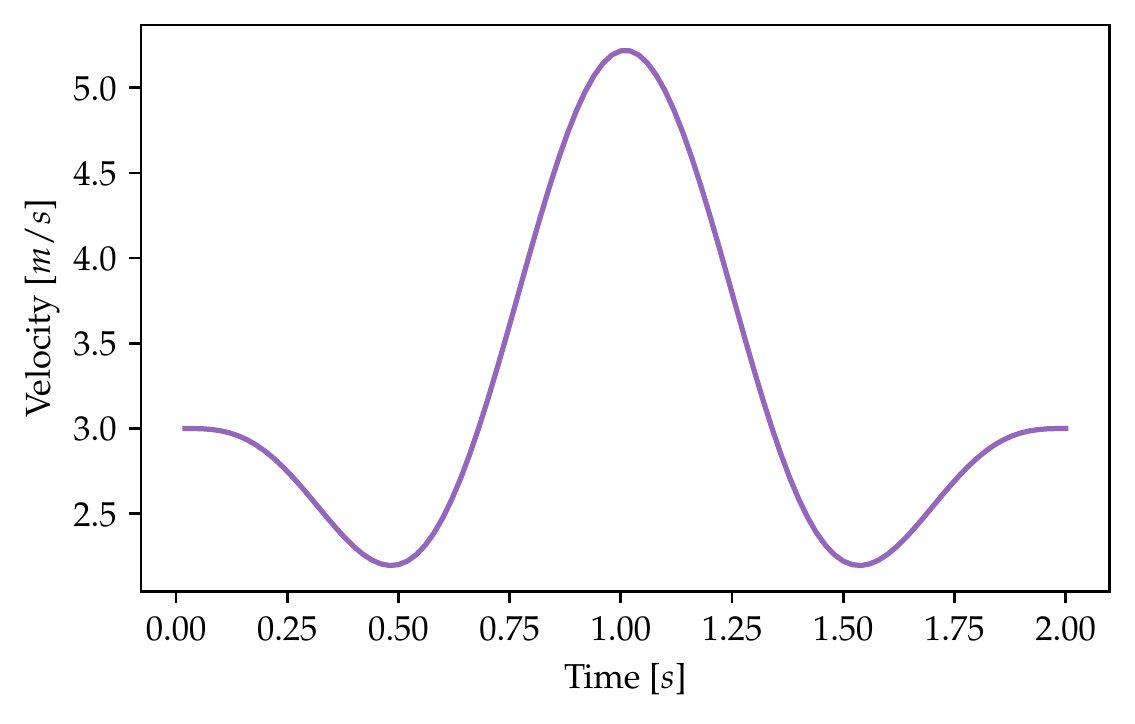}
    \caption{\footnotesize Extreme Operating Gust (EOG) profile used in the variable context experiment. This type of wind gust is modeled by a characteristic ``Mexican hat'' shape. }
    \label{fig:mexican-hat}
\end{figure}

\paragraph{Drone trajectory tracking}
After generating reference trajectories, we employ MPC to obtain control actions over a receding horizon of $10$ time steps. The cost function $J$ is chosen as the mean squared error between each target and real position over time.

\paragraph{Drone stabilization under variable wind profiles}

We also consider stabilizing the system to the origin, i.e. $q^* = \begin{pmatrix}
        x, y, \phi, \dot{x}, \dot{y}, \dot{\phi}
\end{pmatrix} = \begin{pmatrix}
        0, 0, 0, 0, 0, 0
\end{pmatrix}$, under Extreme Operating Gusts (EOG). EOGs can be characterized by a decrease in wind speed, followed by a steep rise, subsequent drop, and final rise back to the original average wind speed \citep{branlard2009wind}:

\begin{equation}
w(t) = \begin{cases}
 \bar{W} - 0.37 W_{gust} \sin{ ( \frac{3 \pi t}{T} ) } (  1 - \cos{ ( \frac{2 \pi t }{T} ) } ) & \text{for}~ t_0 \leq t \leq T\\ 
 \bar{W} & \text{otherwise}
\end{cases}
\label{eq:mexican-hat}
\end{equation}

where $\bar{W}$ is the average wind speed and $W_{gust}$ is the wind gust magnitude. Figure \ref{fig:mexican-hat} shows the wind gust profile used in the drone stabilization experiment. We employ MPC with a receding horizon of $10$ time steps whose cost function $J$ is the mean squared error between each position in the trajectory and $q^*$.

\begin{figure}
    \centering
    \includegraphics[width=0.8\linewidth]{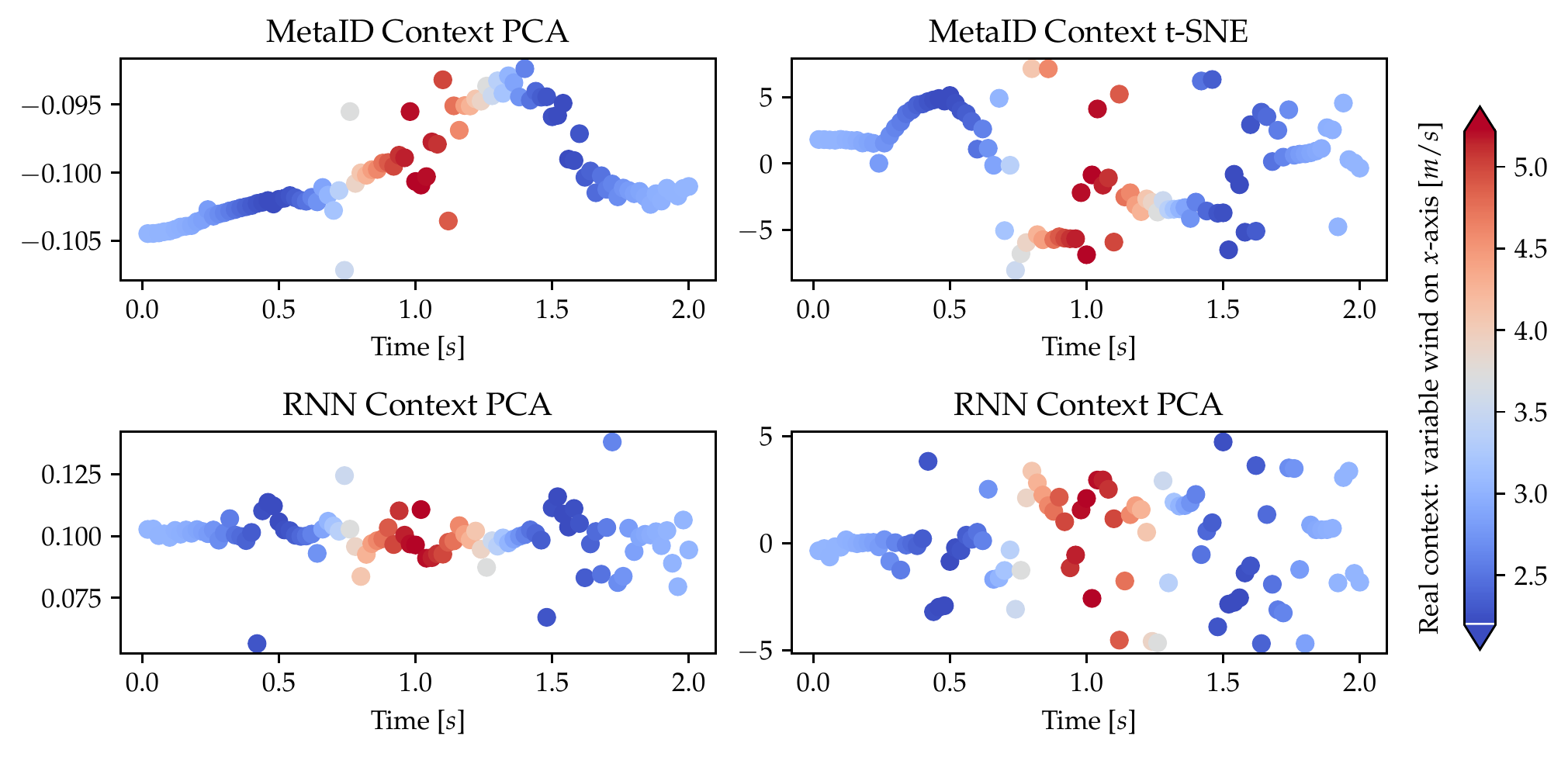}
    \caption{\footnotesize PCA and t-SNE visualizations of \ourmethod{} and the RNN baseline. Even though the latent context is 32--dimensional, \ourmethod{} can reconstruct the roughly smooth dynamics of the wind gust.}
    \label{fig:latent-contexts-drone}
\end{figure}

\paragraph{Latent context visualizations}
We employ PCA \citep{wold1987principal} and t-SNE \citep{van2008visualizing} to visualize the latent context. In particular, we employ the 1--dimensional version in which the latent context of size $32$ is mapped to a value in $\mathbb{R}$. We visualize the operation for each point in time and in \cref{fig:latent-contexts-drone}. We observe that in \model{\ourmethod} the latent context retains more \textit{smoothness} compared to \model{RNN}. We speculate that this can at least partially explain why our proposed method yields better performance in control settings where fast online changes occur.

\subsection{Traffic flow prediction}
\label{appendix:traffic-flow-prediction}

In this section, we detail the model architectures and training process for the traffic flow prediction task.

\paragraph{Model architecture} For the traffic flow prediction \model{\ourmethod} model, we employ the same model architecture of STGCN(Cheb) \cite{yu2017spatio} for $f_\theta$, but modify the input dimension from 15 (i.e., 60-minute observations) to 15+\textcolor{ForestGreen}{32} to accommodate the context as an extra input. 

\paragraph{Training details} We train \model{\ourmethod} with the same hyperparameters of STGCN(Cheb) \cite{yu2017spatio}. To find context, we use 10 inner optimization steps and a step size $\alpha$ of 0.001.

\subsection{Hardware and Software}
\label{appendix:hardware_software}
Experiments were carried out on a machine equipped with an \textsc{AMD Threadripper 2990WX} CPU  with $64$ threads and an \textsc{NVIDIA RTX 3090} graphic card.
Software--wise, we used ${\tt PyTorch}$~\citep{paszke2019pytorch} for deep learning and the ${\tt torchdyn}$ \citep{politorchdyn} library for efficient ODE solvers. 


\section{Additional results}



\subsection{Model-based control results}
\begin{figure}[t]
    \centering
    \includegraphics[width={1\linewidth}]{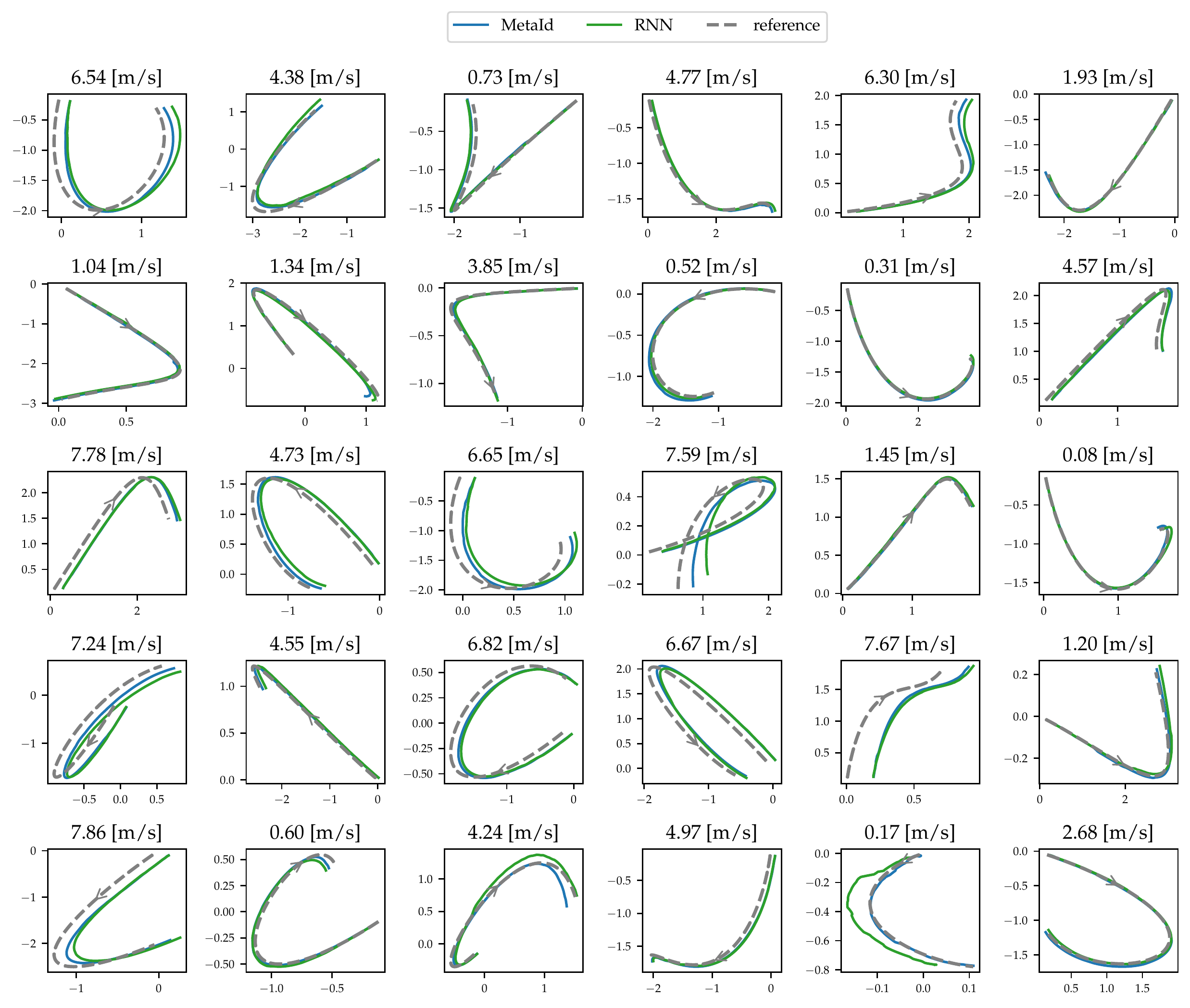}
     \caption{Comparisons between reference and controlled trajectories.}
     \label{fig:ref-track-full}
\vspace{-0.2cm}
\end{figure}

\cref{fig:ref-track-full} visualizes the complete reference tracking results comparing \ourmethod{} with the RNN baseline.

\end{document}